\documentclass[11pt,a4paper]{article}
\usepackage[hyperref]{emnlp2020}
\usepackage{times}
\usepackage{latexsym}

\usepackage{microtype}

\usepackage{algorithm}
\usepackage{graphicx}
\usepackage{subfigure}
\usepackage{booktabs} % for professional tables
\usepackage{multirow}
\usepackage{amssymb}
\usepackage{arydshln} % for hdashline
\usepackage{amsmath}
\usepackage{hyperref}
\usepackage[noend]{algpseudocode}

\newcommand{\tablefont}{\small}

\DeclareMathOperator*{\argmin}{arg\,min}

\newcommand{\boldheader}[1]{\vspace{0.12cm}\noindent\textbf{#1}:}

\aclfinalcopy

\title{Learning to Model and Ignore Dataset Bias\\ with Mixed Capacity Ensembles}

\usepackage{anyfontsize}
\newcommand{\email}{\fontsize{11.2}{12}\selectfont\tt}
\author{
 Christopher Clark\Thanks{Work completed at the University of Washington} \\
 Allen Institute for AI\\
\email{chrisc@allenai.org} \\\And
 Mark Yatskar\Thanks{Work completed at the Allen Institute for AI} \\
 University of Pennsylvania \\
 \email{myatskar@seas.upenn.edu} \\\And
 Luke Zettlemoyer \\
 University of Washington \\
\email{lsz@cs.uw.edu} 
}
\date{}

\begin{document}
\maketitle

\begin{abstract}
Many datasets have been shown to contain incidental correlations created by idiosyncrasies in the data collection process. 
For example, sentence entailment datasets can have spurious word-class correlations if nearly all contradiction sentences contain the word ``not'', and image recognition datasets can have tell-tale object-background correlations if dogs are always indoors. 
In this paper, we propose a method that can automatically detect and ignore these kinds of dataset-specific patterns, which we call dataset biases. 
Our method trains a lower capacity model in an ensemble with a higher capacity model. During training, the lower capacity model learns to capture relatively shallow correlations, which we hypothesize are likely to reflect dataset bias. This frees the higher capacity model to focus on patterns that should generalize better. 
We ensure the models learn non-overlapping approaches by introducing a novel method to make them conditionally independent.
Importantly, our approach does not require the bias to be known in advance.
We evaluate performance on synthetic datasets, and four datasets built to penalize models that exploit known biases on textual entailment, visual question answering, and image recognition tasks. We show improvement in all settings, including a 10 point gain on the visual question answering dataset.

% Our method jointly trains two models in an ensemble, one of which is incentivized to capture relatively shallow correlations, which we hypothesize are likely to reflect dataset bias. This frees the second model to learn patterns that should generalize better. This is accomplished by ensembling a models with different capacities, while using a novel method to enforce conditional independence between the two models.
% To achieve this we 

% We assume that dataset biases can be learned quickly with relatively low capacity models, and present a method that trains such a bias model in an ensemble with a higher capacity model while enforcing conditional independence. Unlike prior  ensemble methods, we do not require the bias to be known in advance; joint learning allows us to discover and factor out the bias and only use the high capacity model at test time.
% shallower and easier to learn than better generalizing patterns by learning to isolate relatively complex and relatively simple patterns in a lower capacity and higher capacity model. 
% The less biased higher-capacity model can then be used alone at test time.
% We assume dataset biases are generally shallower patterns that can be modelled by relatievly low-capacity models.
% Version 1:

\end{abstract}

\section{Introduction}
Modern machine learning algorithms have been able to achieve impressive results on complex tasks such as language comprehension or image understanding. However, recent work has cautioned that this success is often partially due to exploiting incidental correlations that were introduced during dataset creation, and are not fundamental to the target task. Examples include textual entailment models learning the word ``not'' always implies contradiction~\cite{gururangan2018annotation}, visual question answering (VQA) models learning ``2'' is almost always the answer to ``How many'' questions~\cite{jabri2016revisiting}, and question answering models selecting entities that occur near question words irrespective of context~\cite{adversarial_squad}.

We call these kinds of dataset-specific correlations dataset bias. Models that exploit dataset bias can perform well on in-domain data, but will be brittle and perform poorly on out-of-domain or adversarial examples.
Prior work~\cite{clark2019don,hehe,wang2019learning,bahng2019learning} has shown it is possible to prevent models from adopting biased methods, but require the bias to be known and carefully modeled in advance, e.g., by assuming access to a pre-specified classifier that uses only the bias to make predictions. 
In this paper, we present a debiasing method that can achieve similar results, but that automatically learns the bias, removing the need for such dataset-specific knowledge. 

To make this possible, we observe that many known examples of dataset bias involve models learning overly simple patterns~\cite{min2019compositional,mccoy2019right,anand2018blindfold}. This leads us to propose that many dataset biases will be shallower and easier to model than more generalizable patterns.
This reflects the intuition that high-quality models for tasks like language comprehension or image understanding will require some minimum amount of complexity (e.g., a visual question answering model should at least consider the question, image, and ways they might correspond), and therefore shallower approaches are likely to be modelling dataset bias. 

\begin{figure*}
    \centering
    \includegraphics[width=.92\textwidth]{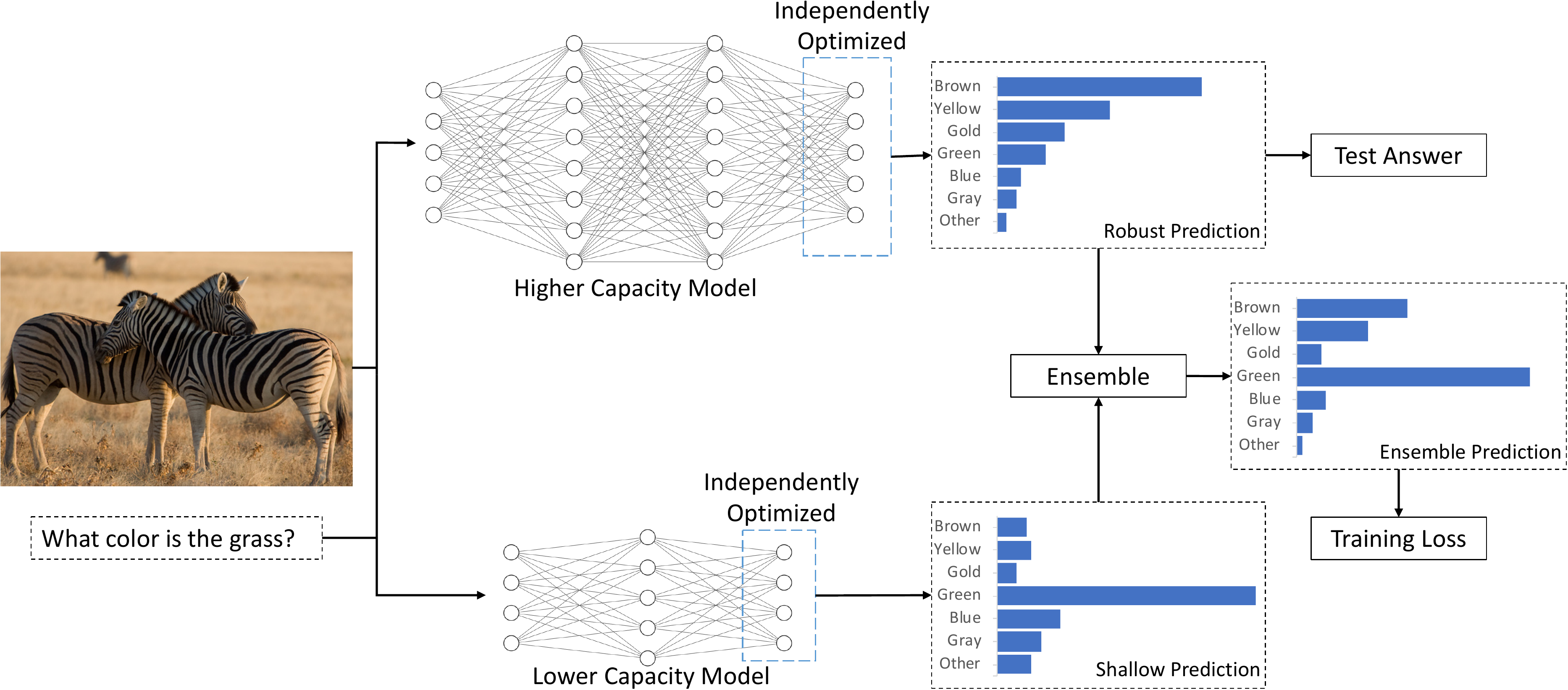}
    \caption{An overview of our method. We train a lower capacity model in an ensemble with a higher capacity model. During training simplistic correlations (e.g., ``grass is usually green'') are captured by the lower capacity model, which frees the higher capacity model to focus on more robust patterns (i.e., matching the question with the image). At test time, the higher capacity model is used alone. We use an independently optimized classifier as the final layer of each model as part of our method to make them conditionally independent (Section~\ref{sect:conditional_independence}). 
    }
    \label{fig:summary_figure}
\end{figure*}

Our method, called Mixed Capacity Ensembling (MCE), follows prior work~\cite{clark2019don,hehe} by training an ensemble of two models, one of which captures bias, and one of which captures other, better generalizing patterns. 
Prior methods required that the model that captures the bias be pre-specified by domain experts. We instead achieve this separation by jointly training both models while encouraging simpler patterns to be isolated into one model and more complex patterns into the other.
In particular, we (1) ensemble a lower capacity model, i.e., a model with fewer parameters, and a higher capacity model, which creates a natural tendency for the higher capacity model to capture more complex patterns, (2) put a small weight on the loss of the lower capacity model so that it is preferentially used to capture simpler patterns that it can model, and (3) enforce conditional independence between the models so that they learn non-overlapping strategies. We show that conditional independence can be achieved by ensuring each classifier makes individually optimal predictions, and we train the ensemble with this constraint using methods from bi-level optimization~\cite{colson2007overview}.

We evaluate our method by training models on datasets with known biases, and then testing them on out-of-domain datasets built to penalize models that learn to use those biases. 
First, we construct a series of synthetic datasets to show our method can adapt to multiple kinds of biases.
Then, we consider three datasets from prior work that test against question-type biases for visual question answering~\cite{vqa2} and hypothesis keyword biases~\cite{bowman2015large,gururangan2018annotation} or lexical overlap biases~\cite{mccoy2019right} for sentence entailment. Finally, we construct an image recognition dataset using Imagenet~\cite{imagenet} that includes a test set of examples with misleading backgrounds (e.g., a fish photographed on dry land) to test our method on background-class biases. 
% We compare against various ablations of our method and several existing techniques from prior work. 
We show improved performance in all settings, in some cases nearly matching the results that can be achieved with an upper-bound that does use knowledge of the bias being targeted.
We release our datasets and code to facilitate future work.\footnote{https://github.com/chrisc36/autobias}

\section{Mixed Capacity Ensembles}
In this section, we present our Mixed Capacity Ensembling method and the motivations behind it. We also discuss an extension to cases where shallow patterns can partially solve the task by eliminating obviously wrong answers.

\subsection{Problem Definition}
Let $\mathcal{X}$ be the domain of the input, $\mathcal{Y} = \{1, 2, \ldots , C\}$ be the space of the labels, and $\mathcal{B}_y$ be the space of probability distributions on $\mathcal{Y}$. 
Assume we have a training dataset of $n$ examples, $\{(x_i, y_i)^n_{i=1}\}$, where $x_i \in \mathcal{X}$ and $y_i \in \mathcal{Y}$ and $x_i, y_i$ are drawn from the joint distribution $P^{train}(x, y)$. 

Our goal is to learn the parameters $\theta_h$ of a differentiable function $f_h$ that returns a vector in $\mathbb{R}^C$ representing a probability distribution over the possible classes, $f_h(\cdot,\theta_h): \mathcal{X} \to \mathcal{B}_y$. For notational simplicity, we will sometimes write $f_h(x_i, \theta_h)$ simply as $f_h(x_i)$.
Our goal is to optimize $\theta_h$ so that $f_h(\cdot,\theta_h)$ will have high accuracy on an out-of-domain test set drawn from $P^{test}(X, Y)$. 

\subsection{Motivation} 
Prior work has generally relied on domain-specific assumptions to make this task possible (e.g., question-only strategies will not generalize well to $P^{test}$ for VQA). Our approach is designed to replace those assumptions with a more domain-general one: that overly simplistic patterns are unlikely to generalize to $P^{test}$. 

While there is no guarantee all examples of dataset bias will fit this criteria, it is commonly true in the growing body of research on dataset bias.
Additionally, we expect complex dataset biases to be less problematic because models will be less prone to fit to them.

% Combining this assumption with the long-held principle in machine learning that overly complex hypotheses are likely to overfit the training data means our goal is to train models that are ``simple, but not too simple."  In other words, while it is often assumed that overly complex hypotheses will overfit individual examples, here we additionally assume overly simple hypotheses will overfit the dataset.

Achieving this through regularization would be challenging since it is not obvious how to penalize the use of simplistic patterns. This motivates our approach of explicitly modeling simplistic hypotheses during training and discarding them during testing.

\subsection{Training an Ensemble}
\label{sect:training_an_ensemble}
Formally, our method introduces a lower capacity model: $f_l(\cdot,\theta_l): \mathcal{X} \to \mathcal{Y}$ and additionally computes a class prior $p_y \in \mathcal{B}_y$ by computing the expected value of $y$ in the training data. We then compute predictions for the ensemble, lower capacity model, and higher capacity model as follows:
%
% \begin{eqnarray*}
\begin{align*} 
\hat{y}_i^{e} &= \mathit{softmax}(\log(f_h(x_i))\! +\! \log(f_l(x_i))\! +\! \log(p_y)) \\
\hat{y}_i^{l} &= \mathit{softmax}(\log(f_l(x_i)) + \log(p_y)) \\ 
\hat{y}_i^{h} &= \mathit{softmax}(\log(f_h(x_i)) + \log(p_y))
\end{align*}
% \end{eqnarray*}

We explicitly factor in $p_y$ so that it can be properly integrated into all three predictions (if the class prior was encoded as part of $f_l$ and $f_h$ it would be double-counted when the two functions are ensembled).
% Where we explicitly model $p_y$ so that it can be properly integrated into all three predictions.
% Failing to do this will result in the ensemble arbitrarily encoding the prior in either $f_l$ or $f_h$, which can lead to sub-optimal results. 
During training the loss is computed as:
\begin{equation}
\label{eq:loss}
\mathit{Loss} = \sum_{i=1}^nL(\hat{y}_i^e, y_i) + wL(\hat{y}_i^l, y_i)
\end{equation}
\noindent
where $L$ is the cross-entropy loss function and $w$ is a hyperparameter. During testing we make predictions using $\hat{y}_i^{h}$.

Following our simplicity assumption, we expect both $f_l$ and $f_h$ to be able to model dataset bias, but due to the additional loss on the output of $f_l$ the ensemble will favor using $f_l$ for that purpose. Additionally, since $f_h$ can better represent more complex patterns, the ensemble will use $f_h$ for that purpose.

\subsection{Adding Conditional Independence}
\label{sect:conditional_independence}
Although this creates a soft incentive for the models to learn different patterns, there is a risk this separation will not be completely clean (e.g., the higher capacity model might partially capture the relatively simple patterns we hope to model with the lower capacity model). To prevent this, we propose to enforce conditional independence between the models, meaning $f_l(X) {\perp\!\!\!\perp} f_h(X) | Y$ for random variables $X$, $Y$ distributed according to $P^{train}$. We do not expect the models to be generally independent, since they will both be predictive of the label. Meeting this requirement while fulfilling the soft incentive created in the previous section pushes the ensemble to isolate simple/complex patterns in the lower/higher capacity models.

Most existing methods for enforcing conditional independence require penalizing some dependency measure between the two models. Here, we propose an alternative that takes advantage of the fact the models are being trained in an ensemble.

\boldheader{Theoretical Motivation} Let $g_h(x_i, \theta_h^g)$ and $g_l(x_i, \theta_l^g)$ be functions that produce feature vectors of size $m$ (meaning $g_h(\cdot, \theta_h^g): \mathcal{X} \to \mathbb{R}^m$). We will again sometimes omit the parameters $\theta_h^g$ and $\theta_l^g$ for notational simplicity.

\newcommand{\vl}{x_l}
\newcommand{\vh}{x_h}
For an example $x$, let $g_l(x) = \vl$ and $g_h(x) = \vh$. Our method is based on the observation that if $\vh$ and $\vl$ are conditionally independent on $y$:
\begin{align} 
\hspace{1em}&\hspace{-2em}P(y|\vh,\vl)  \\
&\propto P(\vh,\vl|y)P(y) \\
&= P(\vh|y)P(\vl|y)P(y) \\
&= \frac{P(y|\vh)P(\vh)}{P(y)} \frac{P(y|\vl)P(\vl)}{P(y)} P(y) \\
&\propto \frac{P(y|\vh)}{P(y)} \frac{P(y|\vl)}{P(y)} P(y)
\end{align}

\noindent
where (3) applies Bayes rule, (4) follows from conditional independence, and (5) applies Bayes rule twice. 
This shows that if $\vh$ and $\vl$ are conditionally independent, then $P(y|\vh,\vl)$ can be computed by ensembling $P(y|\vh)$ and $P(y|\vl)$.

Our key idea is to constrain the model so that $P(y|\vh,\vl)$ must be modelled in this way, i.e., by producing feature vectors $\vl$ and $\vh$, modelling $P(y|\vl)$ and $P(y|\vh)$ individually, and then combining them following equation 6.
Since equation 6 will only be a good model of $P(y|\vh,\vl)$ if $\vh$ and $\vl$ are conditionally independent, this creates a natural incentive to learn conditionally independent feature sets.

Intuitively, conditional independence is achieved because if a particular piece of information relevant to predicting $y$ is present in both $\vh$ and $\vl$, then that information's relationship to $y$ will be captured in both $P(y|\vh)$ and $P(y|\vl)$, and thus will be over-used when these distributions are ensembled. Optimization should remove that information from either $\vh$ and $\vl$ to prevent this over-use. 

\boldheader{Method} We approximate this constraint by training classifiers to model $P(y|\vh)$ and $P(y|\vl)$. Given a classifier $c(\cdot,\theta): \mathbb{R}^m \to \mathcal{B}_y$, we compute:
\begin{align*} 
\theta^{c*}_h &= \argmin_{\theta} \sum^n_{i=1} L(\hat{y}^{\prime}, y_i) \\ 
\hat{y}^{\prime} &= \mathit{softmax}(\log(c(g_h(x_i), \theta) + \log(p_y)) 
\end{align*}
We then set $f_h(x_i) = c(g_h(x_i), \theta^{c*}_h)$, and set $f_c(x_i)$ by following the same procedure to compute $\theta^{c*}_l$ for $g_l$. 
In short, $f_h$ and $f_l$ are now composed of feature extraction functions and classifiers, where the classifiers are fixed to optimally predict $y$ instead of being trained on the end-to-end objective.
As a result, we have $\hat{y}_i^h \approx P(y_i|g_h(x_i))$, in which case $f_h(x_i) \approx P(y|g_h(x_i))/ P(y)$ (because we model the class prior separately from $f_h$ when computing $\hat{y}_i^h$). Therefore the ensemble prediction $\hat{y}^e_i$ is computed following equation 6.

This method does not require adding additional hyperparameters, but it does mean minimizing the loss specified in equation~\ref{eq:loss} becomes a bi-level optimization problem~\cite{colson2007overview} because it contains the subproblem of computing $\theta^{c*}_h$ and $\theta^{c*}_l$. 

\boldheader{Addressing Discontinuities} A complication with this method is that the argmin operations might not be smooth: a small change in $g_h(x_i)$ could dramatically change $\theta^{c*}_h$ and therefore $f_h$. We solve this by regularizing the classifier so that:
$$ \theta^{c*}_h = \argmin_{\theta} \sum^n_{i=1} L(\hat{y}^{\prime}, y_i) + \alpha\Omega(\theta)$$
Where $\Omega(\theta)$ is the L2 norm of the weights. This can also help ensure the optimization problem is well-defined. The hyperparameter $\alpha$ can be tuned by shrinking it until learning is no longer smooth. We find that our method is effective across a range of values for $\alpha$, and we fix $\alpha = 0.002$ for all experiments in this paper. Algorithm~\ref{algo:mce} contains a review of the end-to-end procedure to compute the loss.

% \algdef{SE}[SUBALG]{Indent}{EndIndent}{}{\algorithmicend\ }%
% \algtext*{Indent}
% \algtext*{EndIndent}

\begin{algorithm}
\begin{algorithmic}
\Function{ComputeLoss}{$w,\alpha,x,y,\theta_h,\theta_l,p_y$}
    % \State $z_h = g_h(x, \theta_h)$  \label{algo:forward}
    % \State $z_l = g_l(x, \theta_l)$  \label{algo:forward}
    \State $o_h =$ \Call{Optimize}{$\alpha, g_h(x, \theta_h), y, p_y$}
    \State $o_l =$ \Call{Optimize}{$\alpha, g_l(x, \theta_l), y, p_y$}
    \State $\hat{y}^l = \mathit{softmax}(o_l + p_y)$
    \State $\hat{y}^e = \mathit{softmax}(o_l + o_h + p_y)$
    \State \Return $L(\hat{y}^e, y) + wL(\hat{y}^l, y)$
 \EndFunction
\Function{Optimize}{$\alpha, z, y, p_y$}:
\State $\theta^* = \argmin_{\theta} L(\hat{y}^{\prime}, y) + \alpha\Omega(\theta)$
\State $\hat{y}^{\prime} = \mathit{softmax}(\log(c(z, \theta)) + \log(p_y))$
\State \Return $\log(c(z_h, \theta^*))$
\EndFunction
\end{algorithmic}
\caption{Computing the loss used in MCE. Here we use $x$ and $y$ to refer to the features and labels of a batch of examples.}
\label{algo:mce}
\end{algorithm}

\subsection{Implementation}
\label{sect:implementation}
In practice, we set $m$ to be the number of classes, and build $g_h$ by using the pre-softmax logits from a standard classification model as features. 
We set $c_h$ and $c_l$ to be residual affine functions so that $c_h(g_h(x_i), \theta_h^c) = g_h(x_i)W_h^c + b_h^c + g_h(x_i)$. Using more powerful functions for $c$ could improve the approximation $\hat{y}_i^{l}  \approx P(y|g_h(x_i))$, but linear functions are easier to optimize and sufficient in practice. Residual functions are used so the classifiers can pass-through their input with incurring a regularization penalty.

\boldheader{Optimization}
We minimize the loss through mini-batch gradient descent.
During the forward pass we compute $\theta^{c*}_h$ and $\theta^{c*}_l$ for the minibatch, which essentially requires optimizing a small logistic regression model, and can be done quickly using existing black-box solvers and warm-starting from previous solutions. Once computed, the gradient of the input features with respect to $\theta^{c*}_h$ has a closed-form solution as shown in~\citet{gould2016differentiating}, allowing us to backpropagate gradients through $\theta^{c*}_h$ and $\theta^{c*}_l$ to $g_h$ and $g_l$. 

To help ensure the $\theta^{c*}_h$ and $\theta^{c*}_l$ computed on each minibatch results in a good approximation of $P(y|g_h(x_i))$ and $P(y|g_l(x_i))$, we train with large batches (at least 256 examples) and stratify examples so each class is well represented in each batch. 

\boldheader{Evaluation}
Computing $\theta^{c*}_h$ and $\theta^{c*}_l$ requires using the example's labels, which are not available at test time. Therefore we use values precomputed on a large sample of the training data when classifying unlabelled examples.

\subsection{Answer Candidate Pruning}
\label{sect:answer_pruning}
One risk with this approach is that, while simplistic methods should not be relied upon to entirely solve the task, in some cases they could plausibly be used to eliminate some answer options. For example, given a ``How many'' question in VQA, it can be determined the answer should be a number without looking at the image or the other words. However, our method might factor such simple heuristics into the lower capacity model.

For tasks where this is a possibility, we propose to extract this answer-elimination ability from the lower capacity model through thresholding. At test time we eliminate all answer candidates that the lower capacity model assigns a probability less than some (very conservative) threshold $t$, and select among the remaining candidates using the higher capacity model. This follows the intuition that the simplistic patterns captured by the lower capacity model might still be expected to eliminate very obviously wrong answers in a generalizable way.

\section{Experimental Setup}
We apply our method to datasets that have known biases, and evaluate the trained model on adversarial datasets that were constructed to penalize models that make use of those biases. This is not a perfect evaluation because models might be unfairly penalized for ignoring biases that were unaccounted for when the adversarial dataset was constructed, and therefore still exist in the test set. However, positive results on these cases provide good evidence that our method was at least able to identify the bias the adversarial dataset targeted.

We use standard, high-performing models for the higher capacity models. 
To help ensure the lower capacity models can capture a wide range of possible biases, we pick models that achieve strong results if trained on their own, while still having significantly fewer (i.e., half or less) parameters then the higher capacity model.

For each dataset, we evaluate on an out-of-domain (OOD) and in-domain (ID) test set.
All reported results are averages of ten runs unless otherwise specified.
We provide overviews of the datasets and models used, but leave other training details to the appendix. 

\subsection{Comparisons}
We compare MCE with two ablations, a baseline approach to conditional independence using an adversarial method, and an upper bound that uses domain-specific knowledge of the bias. 

\boldheader{No BP} We train the ensemble without backpropagating through the argmin operators that compute the parameters of the top-level classifiers, $c_l$ and $c_h$. Instead the parameters are treated as constants. This tests whether it is necessary to optimize the model using bi-level optimization methods, or if a more ad-hoc approach would have been sufficient.

\boldheader{No CI} The ensemble is trained without our approach to enforcing conditional independence, meaning it is trained as specified in Section~\ref{sect:training_an_ensemble}.

\boldheader{With Adversary}
The ensemble is trained while replacing our conditional independence method with the `Equality of Odds' adversarial approach from~\citet{zhang2018mitigating}. 
This approach trains linear classifiers that use the log-space output of either the higher or lower capacity model, and a one-hot encoding of the label, to predict the output of the other model. The two classifiers are trained simultaneously with the higher and lower capacity models, which in turn are adversarially trained to increase the loss of those classifiers.
Since we observe it can cause training to diverge, we do not backpropagate gradients from the adversarial loss to the model providing the labels for the classifiers.

\boldheader{Pretrained Bias} Following~\citet{clark2019don}, we construct a bias-only model by training a model that is hand engineered to make predictions using only the target bias (e.g., a hypothesis-only model for MNLI). The high capacity  model is then trained in an ensemble with that pre-trained model.
This method makes use of precise knowledge about the target bias, so we use it as an approximate upper bound on what could be achieved if the target bias was perfectly factored into the lower capacity model. See the appendix for details.

\subsection{Applying Prior Work}
We attempted to apply the methods from~\citet{clark2019don} by having the lower capacity model fill the role of a bias-only model. This means training the lower capacity model on its own, freezing its parameters, and then training the higher capacity model in an ensemble with the pre-trained lower capacity model. However, we found this always leads to poor performance, sometimes to levels close to random chance. The likely cause is that the lower capacity models we use are strong enough to learn far more than just the bias when trained on their own. Therefore the methods from~\citet{clark2019don}, which prevent the higher capacity model from using strategies the lower capacity model learned, leads to the higher capacity model avoiding valid approaches to the task.

\subsection{Same Capacity Sanity Check}
We also sanity checked our method by training MCE using the same model for both the lower and higher capacity models. This consistently leads to poor performance, often worse then training the model without any debiasing method, so we do not show results here. 
We found the lower capacity model often ended up performing as well, or better, then the higher capacity on the in-domain data, which suggests it had learned complex patterns.

\subsection{Hyperparameters}
We use OOD development sets drawn from the same distribution as the OOD test sets for hyper-parameter tuning. As pointed out by prior work~\cite{clark2019don}, this is not an ideal solution since it means some information about the OOD distribution is used during training. As a best-effort attempt to mitigate this issue, we use a single hyperparameter setting for MCE ($w = 0.2$) on all our experiments. Although setting this parameter did require using examples from the OOD dev sets, the fact a single value works well on a diverse set of tasks suggests that our method will generalize to settings where tuning hyperparameters on the OOD test distribution is not possible. Results with optimized hyperparameters are in the appendix. 

The With Adversary baseline also requires hyperparameter tuning. We were unable to find a universally effective setting, so we report results using the best setting found on the OOD dev set. This is indicated by a $^\star$ next to that method.

\begin{table*}
\centering
    \tablefont
\begin{tabular}{lcccccc} \toprule
\multirow{2}{*}{Method} & \multicolumn{2}{c}{Background} & \multicolumn{2}{c}{Patch} & \multicolumn{2}{c}{Split}\\
 & OOD Acc & ID Acc & OOD Acc & ID Acc & OOD Acc & ID Acc\\ \midrule
MCE (Ours) & 81.21 & 93.92 & 74.34 & 86.70 & 92.36 & 93.69 \\
 \hspace{1 mm}No CI & 78.75 & 94.95 & 69.40 & 85.53 & 91.29 & 93.94 \\
 \hspace{1 mm}No BP & 78.39 & 94.34 & 71.93 & 86.67 & 92.35 & 93.78 \\
\hspace{1 mm}With Adversary$^\star$ & 80.57 & 93.37 & 70.31 & 82.78 & 91.24 & 93.43 \\
None & 67.76 & 95.46 & 58.53 & 90.34 & 88.77 & 94.72 \\
\hdashline
Pretrained Bias & 84.59 & 91.17 & 72.96 & 79.72 & 91.27 & 91.61 \\
\bottomrule
\end{tabular}
    \caption{Accuracy on MNIST with synthetic biases when those biases are removed (OOD) or preserved (ID).}
    
    \label{tab:mnist}
\end{table*}

\section{Results}

\subsection{Synthetic MNIST}

We build synthetic datasets by adding a synthetic bias to the MNIST images~\cite{lecun1998gradient}. In particular, we design a superficial image modification for each label, apply the modification that corresponds to each image's label 90\% of the time, and apply a different, randomly selected modification otherwise. OOD sets are built by applying randomly selected modifications.
We use 200 examples for each digit for training, 1000 examples per digit for the OOD dev set, and 1000 per digit for the OOD and ID test sets. Runs are averaged over 100 trials. We use three kinds of modifications in order to demonstrate our method is effective against multiple types of bias.

\textit{Background}: The background of the digit is colored one of 10 colors depending on the label.

\textit{Patch}: The background is divided into a 5x5 grid, the upper left patch is colored to match the label, and the rest of the patches are colored randomly.

\textit{Split}: Following~\citet{feng2019misleading}, the label is encoded in two separate locations in the image. Each label is mapped to a set of color pairs. Then the image is divided into vertical stripes, the center strip is colored randomly, and the other strips are colored using a randomly selected color pair for the example label. To avoid an excessive number of color pairs, we only use the digits 1-8, and map those digits to four super classes.

\boldheader{Higher Capacity Model} We use a model with one convolution layer with 8 7x7 filters and ReLU activation, followed by a 128 dimensional ReLU layer and a softmax predictor layer.

\boldheader{Lower Capacity Model} We use a model with a 128 dimensional ReLU layer then a softmax layer.

\boldheader{Results} 
Table~\ref{tab:mnist} shows the results.
The Patch bias proves to be the hardest, possibly because the patchwork background distracts the model, while the more subtle Split bias is the easiest. Despite using no knowledge of the particular bias being used, our method improves upon training the model naively by at least four points, in two cases slightly out-performing the Pretrained Bias method.
Using the adversary is comparable in some cases, but falls behind on the Patches bias. 
% The lower capacity model is able to get close to 100\% on the train data when pretrained, rendering the Ensemble +H and Bias Product methods ineffective. 

\subsection{VQA}
We evaluate on the VQA-CP v2 dataset~\cite{vqa_cp}, which was constructed by re-splitting VQA 2.0~\cite{vqa2} data into new train and test sets such that the correlations between question types and answers differs between each split. For example, ``white" is the most common answer for questions that start with ``What color is..." in the train set, whereas ``black" is the most common answer for those questions in the test set. 
% Models need to learn to avoid relying on these correlations to do well on the test set.
We hold out 40k examples from the train set to serve as an ID test set, and 40k of the 200k examples in the test set for an OOD dev set.\footnote{Prior work tuned hyper-parameters directly on the test set, but we think its preferable to shrink the test set slightly in order to have a separate OOD dev set} 

Our model follows standard practice of predicting an answer from a set of pre-selected answer candidates. 
Since many of the answers are uncommon, and thus will be poorly represented in individual mini-batches, we cannot apply our conditional independence method out of the box. Instead, we cluster answer candidates by putting the 10 most common answers in individual clusters and the rest in an 11th cluster, and then apply our method while sharing parameters between answers in the same cluster. 
Since the model uses a sigmoid prediction function, we make $g$ a simple two-parameter function that rescales and shifts the input. 

For this dataset, we additionally show results when using the answer candidate pruning method from Section~\ref{sect:answer_pruning} for models where it is applicable. We pick a conservative threshold such that the correct label would be pruned less than 0.1\% of the time on in-domain data.

\boldheader{Higher Capacity Model} We use LXMERT~\cite{tan2019lxmert}, a transformer based model that has been pretrained on image-caption data.

\boldheader{Lower Capacity Model} We make predictions using mid-level representations from the higher capacity model (see the appendix for details). 
% \boldheader{Training} We train for three epochs with a linear decaying learning rate scheme as recommend by the authors.

\boldheader{Results}
Table~\ref{tab:vqa} shows the results.
MCE closes most of the gap between the basic model and the upper bound, suggesting it was able to identify the question-type bias. The baselines underperform MCE by a significant margin, while answer candidate pruning offers a consistent boost.

\begin{table}
    \centering
    \tablefont
\begin{tabular}{lcccc} \toprule 
\multirow{2}{*}{Method} & \multicolumn{2}{c}{OOD Acc} & \multicolumn{2}{c}{ID Acc} \\
 & Acc & w/o AP &  Acc & w/o AP \\ \midrule
MCE (Ours) & 68.44 & 66.10 & 74.03 & 72.72\\
 \hspace{1 mm}No CI & 59.10 & 37.32 & 67.56 & 49.28\\
 \hspace{1 mm}No BP & 61.64 & 47.55 & 67.99 & 55.54\\
\hspace{1 mm}With Adversary$^\star$ & 66.08 & 26.16 & 72.17 & 37.47\\ 
None & 57.65 & - & 76.95 & -\\ 
% Bias Product & 36.32 & 31.98 & 42.62 & 38.40\\ 
% Ensemble +H & 15.71 & 12.13 & 15.36 & 12.42\\
\hdashline
Pretrained Bias & 70.32 & - & 70.78 & -\\ 
\bottomrule
\end{tabular}
    \caption{Results on the VQA-CP OOD test set, and a held out ID test set. We show accuracy with and without answer pruning. Note the ID set is for VQA-CP, and is not comparable to the standard VQA 2.0 dev set.}
    \label{tab:vqa}
\end{table}

\subsection{MNLI}
We evaluate on the MNLI Hard sentence pair classification dataset from~\citet{gururangan2018annotation} and the HANS dataset from~\citet{mccoy2019right}. 

The MNLI Hard dataset is built by training a classifier to predict the target class using only the hypothesis sentence, and then filtering out all examples from the dev set that this classifier is able to classify correctly. Our classifier reaches 54\% accuracy on the matched dev set (compared to 33\% from random guessing) by making use of correlations between certain words and the class (e.g., ``not'' usually implies the class is ``contradiction''). We use 4.4k examples filtered from the MNLI matched dev set as an OOD dev set, and another 4.4k examples filtered from the mismatched dev set as the OOD test set. We use the entire 9.8k mismatched dev set for the ID test set.

The HANS dataset contains 30k examples where both sentences contain similar words, so models that naively classify such sentences as 'entailment' will perform poorly. We do not tune hyper-parameters on HANS, and instead use the same settings as MNLI Hard. 

We evaluate each model on both adversarial sets, so doing well requires models to be simultaneously robust to both the biases being tested for. We build a Pretrained Bias model for hypothesis-only bias, and report the best ensemble result from~\citet{clark2019don} as an upper bound for the HANS dataset.

\boldheader{Higher Capacity Model} We use the pre-trained uncased BERT-Base model~\cite{bert}.

\boldheader{Lower Capacity Model} We use a modified version of the neural bag-of-words model from~\citet{parikh2016decomposable}\footnote{We were unable to get positive results using layer 3 or 6 of the BERT model, possibly because even the lower layers of BERT have a lot of representational power}.

\boldheader{Results}
Table~\ref{tab:mnli} shows the results. BERT-Base is already reasonably robust to the hypothesis-only bias, with only a 4\% gap between the upper bound and the unmodified model, but is more vulnerable to the word-overlap bias in HANS. Our method is able to almost cut the gap in half, with the adversarial approach to conditional independence slightly underperforming MCE. 

\begin{table}
\tablefont
    \centering
\begin{tabular}{lccc} \toprule \\
Method & Hard & HANS &  ID \\ \midrule
MCE (Ours) & 77.58 & 64.43 & 83.28\\
 \hspace{1 mm}No CI & 77.24 & 64.25 & 83.38\\
 \hspace{1 mm}No BP & 76.44 & 62.18 & 83.88\\
\hspace{1 mm}With Adversary$^\star$ & 77.10 & 63.03 & 83.35\\
None & 75.62 & 61.04 & 84.28\\
\hdashline
Pretrained Bias & 79.78 & 62.23 & 76.90\\
\citet{clark2019don} & - & 67.92 & -\\
\bottomrule
\end{tabular}
\caption{Accuracy on two OOD textual entailment datasets (MNLI Hard and HANS) and the mismatched dev set performance (ID) .}
    \label{tab:mnli}
\end{table}

\subsection{ImageNet Animals}
\begin{table}
    \centering
    \tablefont
\begin{tabular}{lcccc} \toprule 
\multirow{2}{*}{Method} & \multicolumn{3}{c}{OOD Acc} & \multicolumn{1}{c}{ID Acc} \\
 & 5\% & 10\% & 20\% & All \\ \midrule
MCE (Ours) & 80.01 & 81.72 & 83.84 & 90.05\\
 \hspace{1 mm}No CI & 80.97 & 82.29 & 84.25 & 89.83\\
\hspace{1 mm}With Adversary$^\star$ & 79.16 & 81.05 & 83.42 & 89.91\\
None & 78.42 & 80.53 & 83.22 & 90.70\\
\hdashline
Pretrained Bias & 87.71 & 87.54 & 86.93 & 78.38\\
\bottomrule
\end{tabular}
    \caption{Results on ImageNet animal recognition on examples where less than  5\%, 10\%, or 20\% of the image-patch classifiers are accurate (OOD), as well as on the entire test set (ID).}
    \label{tab:imagenet}
\end{table}

We build an image recognition dataset that tests
the ability of models to avoid learning background-class correlations. Since training models on ImageNet is computationally expensive, and the large number of classes creates complications when applying our conditional independence approach (i.e., we would have take steps to prevent classes becoming too sparsely represented in each minibatch, as we did for VQA), we build a simplified animal classification dataset by grouping various animal classes into 6 super-classes. See the appendix for details.

% We build a train set with 10k images per class, a dev set with 3k images per class, and a test set with 7k images per class.
We build a train, dev, and test set with 10k, 3k, and 7k images per class respectively.
Similarly to our approach to MNLI, and~\citet{hendrycks2019nae}, we construct OOD datasets by filtering out correct predictions made by a biased classifier. Our biased classifier modifies the ResNet-18 model~\cite{resnet} to build features for 9x9 image patches following BagNet~\cite{bagnet}. We then train a classifier to predict the class using those features. Images where the classifier was unable to guess the correct class for most of the image patches are assumed to have a misleading background and are used for the OOD test set. Examples are shown in the appendix.

\boldheader{Higher Capacity Model} We use ResNet-18~\cite{resnet}.

\boldheader{Lower Capacity Model} We branch the higher capacity model by adding a 1x1 convolution with 256 filters after the first residual block, followed by max-pooling and a 256 dimensional ReLU layer.\footnote{We found positive, but slightly weaker results using features after the second residual block, and negative results when using features after the third residual block} This model gets approximately 81\% on our dev set if trained alone.

\boldheader{Results} 
Table~\ref{tab:imagenet} shows the results. 
This dataset proves to be challenging; MCE provides a boost over naively training the model, but is still significantly below our estimated upper bound. Our conditional independence method reduces performance here, and training fails to converge for the no-backpropagation ablation, so results for that method are not shown. This appears to be because the model is able to reach nearly 100\% accuracy on the training data which causes the argmin operations to become degenerate. 

\subsection{Discussion}
\label{sect:discussion}
Overall, we are able to improve out-of-domain performance in all settings even though our method does not use dataset-specific information about the target bias. Our conditional independence method generally improved performance while the adversarial baseline, despite getting the benefit of per-dataset hyperparameter tuning, was less effective. 

MCE decreases ID performance, which is expected since the bias is helpful on in-domain data and the goal of our method is to remove it. However the decrease is often much less than for the upper bound (e.g., on MNLI MCE is 2 points behind on the OOD test set, but 6 points ahead on the ID test set). A possible cause is that the bias is only being partially factored out. Improving OOD performance without losing much ID performance might also suggest MCE is helping improve the higher capacity model in general.
Better understanding and making use of this phenomenon is an interesting avenue for future work. 

We find computing the argmin operations adds a moderate computational overhead, requiring about 2x the train time for the ResNet-18 ensemble and about 1.3x the time for the larger BERT and LXMERT ensembles (performance during evaluation is identical). Our implementation performs the optimization on the CPU; a GPU based optimizer might reduce this overhead.

\section{Related Work}
Prior work has shown that, given precise knowledge of a dataset bias, it is possible to train a debaised model. For example, if it is known particular intermediate representations in a network could only contain features that are shallowly indicative of the class due to bias (e.g., a local patch in an image), adversarial networks can be used~\cite{belinkov2019adversarial,grand2019adversarial,wang2019learning,cadene2019rubi}. 
Alternatively, given a model that is so tightly constrained it can only utilize dataset bias (e.g., a question only model for VQA), REBI~\cite{bahng2019learning} employs a conditional independence penalty between that model and a debaised model, HEX~\cite{wang2019learning} constructs a feature space that is orthogonal to the one learned by the bias model, and~\citet{clark2019don} and~\citet{hehe} pre-train the bias model and then train a debiased model in an ensemble with it. Our approach also makes use of ensembling, and the idea of requiring conditional independence between the models. However, our method identifies biased strategies during training instead of requiring them to be pre-specified.

Additional work has used multi-label annotations~\cite{singh2020don}, pixel-level annotations~\cite{hendricks2018women}, or other annotated features~\cite{kim2019learning} to help train debiased models, although these kinds of annotations will not always be available.

There are also debiasing strategies that identify hard examples in the dataset and re-train the model to focus on those examples~\cite{yaghoobzadeh2019robust,li2019repair,le2020adversarial}. This approach reflects a similar intuition that simplicity is connected to dataset bias, although our method is able to explicitly model the bias and does not assume a pool of bias-free examples exist within the training data. 

Another debiasing approach is to use pretraining~\cite{lewis2018generative,carlucci2019domain} or carefully designed model architectures~\cite{vqa_cp,zhang2016yin,carlucci2019domain} to make models more prone to focus on semantic content. We expect these methods to be complementary to our work; for instance we show we can improve the performance of the extensively pretrained BERT and LXMERT models.

A related task is domain generalization, where the goal is to generalize to a unseen test domain given multiple training datasets from different domains~\cite{muandet2013domain}. 
% Most existing work on this field has been on image recognition, using datasets including PACS~\cite{pacs} or Office datasets~\cite{office_dataset}. 
Most domain generalization methods learn a data representation that is invariant between domains through the use of domain-adversarial classifiers~\cite{ganin2016domain,li2018deep}, ensembling domain-specific and domain-invariant representations~\cite{bousmalis2016domain,ding2017deep} or other means~\cite{arjovsky2019invariant,li2018domain,xu2014exploiting,ghifary2015domain}. Our approach is similar to the ensembling methods in that we explicitly model generalizable and non-generalizable patterns, but does not require multiple training datasets.

% Could delete?
%Improving model robustness to out-of-domain examples built by applying adversarial perturbations to existing examples has also been the subject of recent study~\cite{xie2019feature, goodfellow2014explaining}. See~\citet{carlini2019evaluating} for an overview. Our method is built to defend against larger, more semantic domain shifts. 

\section{Conclusion}
We have presented a method for improving out-of-domain performance by detecting and avoiding overly simple patterns in the training data. Our method trains an ensemble of a lower capacity model and a higher capacity model in a way that encourages conditional independence given the label, and then uses the higher capacity model alone at test time.
Experiments show this approach successfully prevents the higher capacity model from adopting known biases on several real-world datasets, including achieving a 10-point gain on an out-of-domain VQA dataset.

\section*{Acknowledgements}
This work was supported in part by the ARO
(ARO-W911NF-16-1-0121) and the NSF (IIS1252835, IIS-1562364). We thank Kevin Clark, Jungo Kasai, and Tim Dettmers as well as the anonymous reviewers for their feedback on this document, and the members of the UW NLP group for helpful conversations and comments on this work.

\clearpage 

\bibliographystyle{acl_natbib}
\bibliography{main.bib}

\clearpage 

\appendix
\setcounter{figure}{0}
\setcounter{table}{0}
\setcounter{footnote}{0}

\section{ImageNet Animals}
\label{appendix:imagenet-animals}
\begin{figure*}
    \centering
    \includegraphics[width=.92\textwidth]{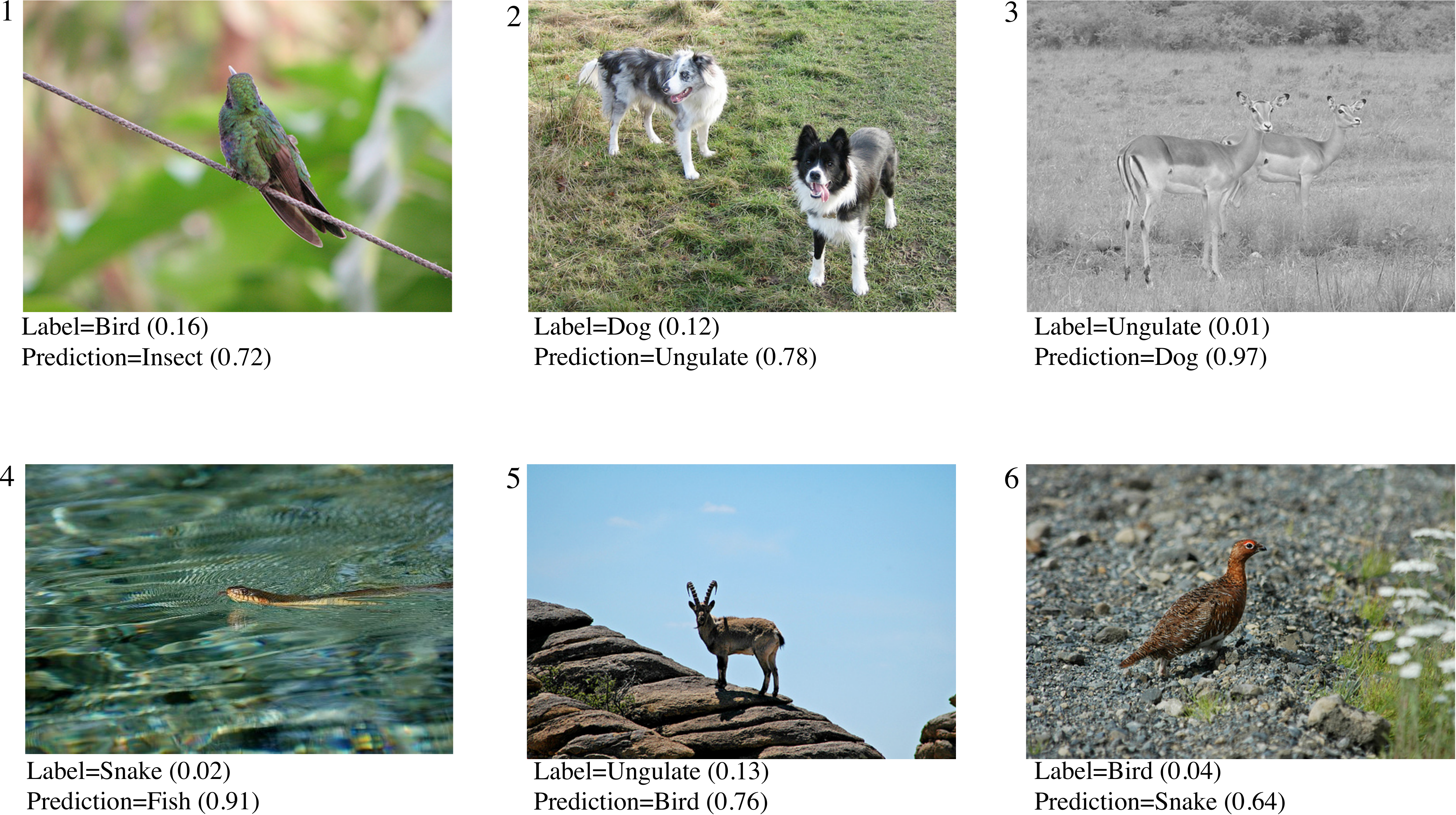}
    \caption{Qualitative examples from ImageNet Animals where most of the image-patch classifiers were incorrect. We show images paired with the gold label and the most common prediction made by the image-patch classifiers, with the percent of image-patch classifiers that predicted those labels in parentheses. Errors are often caused by the patch classifiers associating features of the background with the class. In particular, because (1) twigs and leaves are associated with insects, (2) grassy fields are associated with ungulates (e.i., hooved mammals), (3) black and white photos are associated with dogs due to the commonality of black-and-white dog photos in the training data, (4) water is associated with fish, (5) open sky is associated with birds, and (6) close-ups of the ground are associated with snakes.
    }
    \label{fig:imagenet_qualitative}
\end{figure*}

Here, we describe how ImageNet Animals was built in more details. We built super-classes by taking advantage of the hierarchical structure of ImageNet~\cite{imagenet} and its correspondence to WordNet~\cite{fellbaum2012wordnet}. We selected 6 wordnet synsets that have at least 20 hyponyms that exist in ImageNet: fish.n.01, insect.n.01, dog.n.01, bird.n.01, ungulate.n.01, and snake.n.01. Each synset is used as a class. We gather examples for each class by sampling images randomly from all its hyponyms in the ImageNet training data. Examples of what kinds of images were considered OOD are shown in Figure~\ref{fig:imagenet_qualitative}.

\section{Training Details}
In this section, we give additional details about the models used, and specify the learning rates, batch sizes, and other optimization details. All models were trained with 1 to 4 GeForce RTX 2080 Ti GPUs.

\subsection{Synthetic MNIST}
We trained all models using a learning rate of 0.01, batch size of 1000, and momentum of 0.9. Models were trained until loss on the training data stopped decreasing, and were trained on a single GPU. For regularization, dropout was applied before the last layer of both the higher capacity and lower capacity model with a rate of 0.5.

\subsection{VQA}
Models were trained using Adam~\cite{adam} and a linearly decaying learning rate, following the default settings for LXMERT~\cite{tan2019lxmert}\footnote{https://github.com/airsplay/lxmert}, except that we use a batch size of 512 instead of 32, and train for 3 epochs instead of 4. 
We found these changes to the optimization procedure increased performance by a significant margin; on the standard VQA 2.0 dataset our implementation achieves a dev set accuracy of 73.4 with them and 70.1 without them (compared to 69.9 reported by~\citet{tan2019lxmert}). 
Models were trained with 4 GPUs, and were trained in floating-point 16 mode using Apex.\footnote{https://github.com/NVIDIA/apex}

 We also upsample examples for each answer cluster to ensure each cluster consists about 5\% of the training data, while putting proportional weights on the classes so that the weighted answer distribution matches the original training data. This ensures each mini-batch can contain a reasonable number of examples for each class.

\subsubsection{Lower Capacity Model}
In more detail, the lower capacity model takes the intermediate question-only and image-only representations from the higher capacity model, concatenates these vectors and their elementwise product, and passes the result to a 2048 ReLU layer, followed by a linear predictor layer. This model reaches 67\% on the VQA 2.0 dev set when trained alone.

\subsection{MNLI}
We use the default BERT optimizer~\cite{bert}, which is known to be effective for MNLI, but with a batch size of 256. We found increasing the batch size did not change performance on the matched MNLI dev set. Since the lower capacity model is not pretrained, we use a learning rate of 1e-3 without linear decay instead of 5e-5 for its parameters. Models were also trained with 4 GPUs, and were trained in floating-point 16 mode using Apex.

\subsubsection{Lower Capacity Model}
Here, we specify the lower capacity model in more detail. The model is a simplified version of the model from~\citet{parikh2016decomposable} and has the following stages:

\textbf{Embed:} Embed the words using a character CNN, following what was done by \citet{seo2016bidirectional}, and the fasttext crawl word embeddings~\cite{fasttext_word_vectors}, then apply a 200 dimensional ReLU layer.

\textbf{Co-Attention}: Compute an attention matrix using the formulation from~\citet{seo2016bidirectional}, and use it to compute a context vector for each premise word~\cite{bahdanau2014neural}. Then build an augmented vector for each premise word by concatenating the word's embedding, the context vector, and the elementwise product of the two.
Augmented vectors for the hypothesis are built the same way using the transpose of the attention matrix.

\textbf{Pool}: Feed the augmented vectors into another 200 dimensional ReLU layer, and max-pool the results. 
The max-pooled vectors from the premise and hypothesis are concatenated and fed into a softmax layer with three outputs to compute class probabilities.
\\
\\
We apply dropout at a rate of 0.1 before all fully connected layers.

\subsection{ImageNet Animals}
Models are trained with a learning rate of 0.02, a momentum of 0.9 and batch size of 512 for 76 epochs. The learning rate is decreased to 0.006 for the last 10 epochs. Models were trained with 4 GPUs.

\section{Pretrained Bias Upper Bounds}
In this section, we describe the Pretrained Bias approach in more detail. 
The general method is to train a bias-only model that captures the target bias, and then use it to train a de-biased model using either the Bias Product or Entropy +H method from~\citet{clark2019don}.

\subsection{Synthetic MNIST}
We train a bias-only model by training a model to predict the label using a one-hot encoding of which modification was applied to the image. This model is then used with the Bias Product method, since that method is expected to be the best choice when the bias-only model perfectly captures a conditionally independent bias.

\subsection{VQA}
We follow~\citet{clark2019don} by training a bias-only model that predicts the answer using a one-hot encoding of the question-type, and using the Ensemble +H method. The entropy penalty was set to 0.24 using the OOD dev set.

\subsection{MNLI Hard}
We use the hypothesis-only model that was used to construct the dataset splits as a bias-only model, and apply the Ensemble +H method. The value of the entropy penalty was selected to be 0.1 using the OOD dev set.

\subsection{ImageNet Animals}
We build a bias-only model by computing the expected class prediction across all image-patch classifiers. Since it is possible for this prediction to be zero for some classes, we additionally smooth the distribution by adding $\log(1 + e^{\alpha})$ to the probabilities and re-normalizing, where $\alpha$ is a learned smoothing parameter. This model is then used with the Entropy +H method, the value of the entropy penalty was again set to 0.1 based on the OOD dev set.

\section{Results with Optimized Hyperparameters}
In this section, we show results when using optimized hyperparameters for MCE, i.e., when $w$ is tuned. 

\subsection{Synthetic MNIST}
Table~\ref{tab:opt_mnist} shows the results. On these datasets increasing $w$ can sometimes lead to small performance improvements for MCE and its ablations.

\begin{table*}
\tablefont
\begin{tabular}{lccccccccc} \toprule
\multirow{2}{*}{Method} & \multicolumn{3}{c}{Background} & \multicolumn{3}{c}{Patch} & \multicolumn{3}{c}{Split}\\
 & H & OOD Acc & ID Acc & H & OOD Acc & ID Acc & H & OOD Acc & ID Acc\\ \midrule
MCE (Ours)$^\star$ & 0.5 & 82.14 & 93.99 & 0.8 & 75.22 & 87.57 & 1.0 & 93.06 & 94.01 \\
\hspace{1 mm}No CI$^\star$ & 0.15 & 78.83 & 94.92 & 1.0 & 70.25 & 82.19 & 0.05 & 91.47 & 93.86 \\
 \hspace{1 mm}No BP$^\star$ & 1.0 & 80.13 & 93.93 & 0.2 & 71.93 & 86.67 & 1.0 & 92.52 & 93.89 \\
\hspace{1 mm}With Adversary$^\star$ & 0.05/0.08 & 80.57 & 93.37 & 0.7/0.01 & 70.31 & 82.78 & 0.02/0.01 & 91.21 & 93.56 \\
\bottomrule
\end{tabular}
\caption{Results on Synthetic MNIST with hyperparameter tuning. The hyperparameter(s) used (either $w$ for MCE or $w$ and the adversary weight for With Adversary) are shown in the H columns.}
\label{tab:opt_mnist}
\end{table*}

\subsection{VQA}
Table~\ref{tab:opt_vqa} shows the results. For VQA $w=0.2$ is the optimal setting for MCE, although decreasing $w$ leads to performance gains for the ablated versions.

\begin{table*}
\tablefont
\centering
\begin{tabular}{lcccccc} \toprule
\multirow{2}{*}{Method} & \multirow{2}{*}{H} & \multicolumn{2}{c}{OOD Acc} & \multicolumn{2}{c}{ID Acc} \\
 & & Acc & w/o AP &  Acc & w/o AP \\ \midrule
 
MCE (Ours)$^\star$ & 0.2 & 68.44 & 66.10 & 74.03 & 72.72\\ 
 \hspace{1 mm}No CI$^\star$ & 0.1 & 65.40 & 55.77 & 74.38 & 66.01\\ 
 \hspace{1 mm}No BP$^\star$ & 0.1 & 63.65 & 52.43 & 70.76 & 60.82\\ 
\hspace{1 mm}With Adversary$^\star$ & 0.2/0.007 & 66.08 & 26.16 & 72.17 & 37.47\\ 
% \hdashline
% Pretrained Bias & 0.24 & 70.32 & 0.00 & 70.78 & 0.00\\ 
\bottomrule
\end{tabular}
\caption{Results on VQA-CP with hyperparameter tuning. The hyperparameters used are shown in the H column.}
\label{tab:opt_vqa}
\end{table*}

\subsection{MNLI}
Table~\ref{tab:mnli_opt} shows the results. For MNLI $w=0.2$ is consistently the best setting, so the results are unchanged.

\begin{table*}
    \tablefont
    \centering
    \begin{tabular}{lcccc} \toprule
Method & H & Hard & HANS & ID \\ \midrule 
MCE (Ours)$^\star$ & 0.2 & 77.58 & 64.43 & 83.28\\
 \hspace{1 mm}No CI$^\star$ & 0.2 & 77.24 & 64.25 & 83.38\\
 \hspace{1 mm}No BP$^\star$ & 0.2 & 76.44 & 62.18 & 83.88\\
\hspace{1 mm}With Adversary$^\star$ & 0.3/0.16 & 77.10 & 63.03 & 83.35\\
% \hdashline
% Pretrained Bias & 0.1 & 79.85 & 77.12\\ 
\bottomrule
\end{tabular}
\caption{Results on MNLI with hyperparameter tuning.}
\label{tab:mnli_opt}
\end{table*}

\subsection{ImageNet Animals}
Table~\ref{tab:imagenet_opt} shows the results. Here decreasing $w$ improved performance of MCE, but maintaining $w$ at 0.2 was best for the No CI method.

\begin{table*}[]
    \tablefont
    \centering
\begin{tabular}{lccccc} \toprule \\ 
Method & H & 0.05 & 0.1 & 0.2 & All \\ \midrule 
MCE (Ours)$^\star$ & 0.1 & 80.47 & 81.85 & 83.92 & 90.09\\ 
 \hspace{1 mm}No CI$^\star$ & 0.2 & 81.02 & 82.46 & 84.51 & 89.97\\ 
\hspace{1 mm}With Adversary$^\star$ & 0.3/0.005 & 80.33 & 81.77 & 83.79 & 89.92\\ 
\bottomrule
\end{tabular}
\caption{Results on ImageNet Animals with hyperparameter tuning.}
 \label{tab:imagenet_opt}
\end{table*}

\end{document}